\documentclass[11pt,a4paper]{article}
\usepackage[hyperref]{emnlp2020}
\usepackage{times}
\usepackage{latexsym}
\usepackage{CJKutf8}
\usepackage{amsfonts}
\usepackage{amsmath}
\usepackage{soul, color}
\usepackage{tikz}
\usepackage{booktabs}
\usepackage{bchart}
\usepackage{pgfplots}
\usepackage{multirow}
\usepackage{pgfplots}
\usepackage{array}
\usepackage{titling}
\usepackage{authblk}
\usetikzlibrary{pgfplots.groupplots}
\usetikzlibrary{arrows,shapes,chains, calc, fit, decorations.markings}

\usepackage{microtype}

\aclfinalcopy % Uncomment this line for the final submission

\DeclareFontFamily{OT1}{pzc}{}
\DeclareFontShape{OT1}{pzc}{m}{it}{<-> s * [1.10] pzcmi7t}{}
\DeclareMathAlphabet{\mathpzc}{OT1}{pzc}{m}{it}

\renewcommand*{\Affilfont}{\normalsize\normalfont}
\newsavebox\affbox

\title{\textbf{Focus-Constrained Attention Mechanism for CVAE-based Response Generation}}

\makeatletter
\newcommand\email[2][]%
   {\newaffiltrue\let\AB@blk@and\AB@pand
      \if\relax#1\relax\def\AB@note{\AB@thenote}\else\def\AB@note{\relax}%
        \setcounter{Maxaffil}{0}\fi
      \begingroup
        \let\protect\@unexpandable@protect
        \def\thanks{\protect\thanks}\def\footnote{\protect\footnote}%
        \@temptokena=\expandafter{\AB@authors}%
        {\def\\{\protect\\\protect\Affilfont}\xdef\AB@temp{#2}}%
         \xdef\AB@authors{\the\@temptokena\AB@las\AB@au@str
         \protect\\[\affilsep]\protect\Affilfont\AB@temp}%
         \gdef\AB@las{}\gdef\AB@au@str{}%
        {\def\\{, \ignorespaces}\xdef\AB@temp{#2}}%
        \@temptokena=\expandafter{\AB@affillist}%
        \xdef\AB@affillist{\the\@temptokena \AB@affilsep
          \AB@affilnote{}\protect\Affilfont\AB@temp}%
      \endgroup
       \let\AB@affilsep\AB@affilsepx
}
\makeatother

\author[1]{Zhi Cui}
\author[2]{Yanran Li}
\author[1]{Jiayi Zhang}
\author[1]{Jianwei Cui}
\author[1]{Chen Wei}
\author[1]{Bin Wang}
\affil[1]{Xiaomi AI Lab}
\affil[2]{The Hong Kong Polytechnic University}
\email{\texttt{\{cuizhi,zhangjiayi3,cuijianwei,weichen,wangbin11\}@xiaomi.com}}
\email{\texttt{csyli@comp.polyu.edu.hk}}
\date{}  

\begin{document}
\begin{CJK*}{UTF8}{gbsn}
\maketitle
\begin{abstract}

To model diverse responses for a given post, one promising way is to introduce a latent variable into Seq2Seq models. The latent variable is supposed to capture the discourse-level information and encourage the informativeness of target responses. However, such discourse-level information is often too coarse for the decoder to be utilized. To tackle it, our idea is to transform the coarse-grained discourse-level information into fine-grained word-level information. Specifically, we firstly measure the semantic concentration of corresponding target response on the post words by introducing a fine-grained \emph{focus} signal. 
Then, we propose a focus-constrained attention mechanism to take full advantage of \emph{focus} in well aligning the input to the target response. The experimental results demonstrate that by exploiting the fine-grained signal, our model can generate more diverse and informative responses compared with several state-of-the-art models.\footnote{https://github.com/cuizhi555/Focus-Constrained-Attention-Mechanism-for-CVAE-based-Response-Generation.}
\end{abstract}

\section{Introduction}
\label{sec:section1}
Nowadays, developing intelligent open-domain conversational systems has become an active research field~\cite{perez2011conversational,shum2018eliza}. Compared with rule-based and retrieval-based methods, neural generative models have attracted increasing attention because they do not need extensive feature engineering and have achieved promising results recently with large-scale conversational data \cite{vinyals2015neural,sordoni2015neural,shang2015neural}. 

Typically, neural generative models are trained to learn the post-response mappings based on the Seq2Seq architecture using maximum likelihood~(MLE) training objective. This kind of objective induces the model to treat the post-response relationship as one-to-one mappings. However, the conversations in the real world often embodies one-to-many relationships, where a post is often associated with multiple valid responses~\cite{zhou2017mechanism}. 
Due to this discrepancy, standard Seq2Seq models tend to generate high-frequency but trivial responses such as ``\textit{I don't know}" or ``\textit{I'm ok}"~\cite{li2016diversity}.%, the so-called ``generic response'' issue.

To address this issue, one promising research line resorts to Conditional Variational Autoencoder (CVAE), which introduces a latent variable to Seq2Seq models through variational learning. The latent variable is supposed to capture the discourse-level semantics of target response and in turn encourage the response informativeness. Recent literature along this line attempted to improve the model performance by putting extra control on the latent variable~\cite{zhao2017learning,gu2018dialogwae,gao2019discrete}. Despite the control, these methods still relied on the discourse-level latent variable, which is too coarse for the decoders to mine sufficient guiding signals at each generation step. As a result, these variational models 
are observed to ignore the latent variable~\cite{zhao2017learning,gu2018dialogwae,gao2019discrete} and to generate semantically irrelevant or grammatically disfluent responses~\cite{qiu2019training}.

In this paper, we propose a novel CVAE-based model, which exploits fine-grained word-level information for diverse response generation. Firstly, we transform the discourse-level information into word-level signals, i.e., \emph{focus}. By attending the latent variable to the post words, the \emph{focus} weight measures the response's correlation with the post words. The higher the weight, the semantics is more likely to concentrate on the corresponding word. To 
utilize the \emph{focus}, we develop a focus-constrained attention mechanism which better aligns the post words with the response according to the fine-grained signals. In this way, the model is able to produce a semantically different response directed by a different \emph{focus}.

Our contributions can be summarized as three folds: 1) We propose a novel CVAE-based model for diverse response generation, by directing the decoder with fine-grained information. 2) We introduce \emph{focus} to represent the fine-grained information, and propose a focus-constrained attention mechanism to make full use of it. 3). Experimental results demonstrate our model outperforms several state-of-the-art models in terms of response's diversity as well as appropriateness.

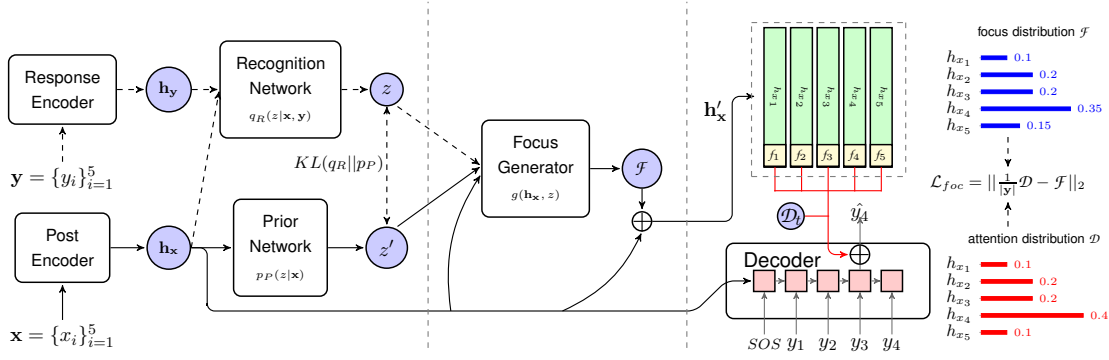
\begin{figure*}
\centering
\resizebox{2.0\columnwidth}{!}{
\begin{tikzpicture}[
  font=\sffamily,
  every matrix/.style={ampersand replacement=\&},
  strip/.style={rectangle, draw,thick,fill=green!20, minimum height=0.41cm, minimum width=0.41cm},
  sstrip/.style={draw,thick,fill=yellow!20, minimum size=0.41cm},
  brick/.style={draw, thick,fill=red!20, minimum size=0.4cm},
  source/.style={draw,thick,rounded corners,inner sep=.3cm},
  ssource/.style={draw, rectangle, thick,rounded corners,inner sep=.5cm, label={[shift={(0ex,-3.5ex)}]north :#1}},
  sssource/.style={draw, rectangle, thick,rounded corners,inner sep=.5cm, label={[shift={(-5ex,-3.5ex)}]north :#1}},
  process/.style={draw,thick,circle,fill=blue!20,minimum size=0.3cm},
  concat/.style={draw,thick,circle},
  sink/.style={source,fill=green!20},
  datastore/.style={draw,very thick,shape=datastore,inner sep=.3cm},
  dots/.style={dashed,gray,scale=2,>=stealth',shorten >=1pt,semithick,font=\sffamily\footnotesize},
  to/.style={->,>=stealth',shorten >=1pt,semithick,font=\sffamily\footnotesize, rounded corners,},
  sto/.style={->,>=stealth',gray!120,font=\sffamily\footnotesize},
  line/.style={>=stealth',shorten >=1pt,semithick,font=\sffamily\footnotesize},
  every node/.style={align=center}
  ]
  
  % Position the nodes using a matrix layout
  \matrix{
    
%    \node[brick ](r1) {};
%    \node[brick, xshift=0.6cm] (r2) {};
%    \node[brick, xshift=-0.6cm] (r3) {};
%    \node[brick, xshift=1.2cm ] (r4) {};
%    \node[brick, xshift=-1.2cm] (r5) {}; 
%    \node[ssource=Response Encoder, fit=(r1)(r2)(r3)(r4)(r5)] (r_encoder) {};
    \node[source] (r_encoder) {\small Response \\ \small Encoder};
     
%    \node[brick, below of=r1, node distance=3cm] (p1) {};
%    \node[brick, below of=r2, node distance=3cm] (p2) {};
%    \node[brick, below of=r3, node distance=3cm] (p3) {};
%    \node[brick, below of=r4, node distance=3cm] (p4) {};
%    \node[brick, below of=r5, node distance=3cm] (p5) {}; 
%    \node[ssource=Post Encoder, fit=(p1)(p2)(p3)(p4)(p5),below of=r_encoder, node distance=3cm] (p_encoder) {};

    \node[source, below of=r_encoder, node distance=3cm] (p_encoder) {\small Post \\  \small Encoder};
    
    \node[process, right of=p_encoder, node distance=2cm] (h_x) {\small $\mathbf{h_x}$}; 
    \node[process, right of=r_encoder, node distance=2cm] (h_y) {\small $\mathbf{h_y}$};
    
    \node[source, right of=h_x, node distance=2.1cm] (p_network) {\small Prior \\ \small Network \\ \tiny $p_{P}(z|\mathbf{x})$};
    \node[source, right of=h_y, node distance=2.1cm] (r_network) {\small Recognition \\ \small Network \\ \tiny $q_{R}(z|\mathbf{x}, \mathbf{y})$}; 
    \node[process, right of=p_network, node distance=2cm] (z_dot) {$z'$};
    \node[process, right of=r_network, node distance=2cm] (z) {$z$};
    \node[source, right of=z, node distance=2.8cm, yshift=-1.5cm] (f_generation) {\small Focus \\ \small Generator \\ \tiny $g(\mathbf{h_x}, z)$};
    \node[process, right of=f_generation, node distance=2cm] (f) {\small $\mathpzc{F}$}; 
    
    \node[process, right of=f_generation, node distance=4.8cm, yshift=-0.9cm,minimum size=0.5cm] (d) {};

    \node[concat, below of=f, node distance=1cm] (concat) {};

  \node[strip, right of=f, node distance=3.5cm, yshift=1.35cm] (h1) { \\ \\ \\ \\ \\ };
  \node[sstrip, below of=h1, node distance=1.1cm] (f1) {};
  \node[strip, right of=f, node distance=3.5cm, yshift=1.35cm, xshift=-0.5cm] (h2) { \\ \\ \\ \\ \\ };
  \node[sstrip, below of=h2, node distance=1.1cm] (f2) {};
  \node[strip, right of=f, node distance=3.5cm, yshift=1.35cm, xshift=0.5cm] (h3) { \\ \\ \\ \\ \\ };
  \node[sstrip, below of=h3, node distance=1.1cm] (f3) {};
  \node[strip, right of=f, node distance=3.5cm, yshift=1.35cm, xshift=-1.0cm] (h4) { \\ \\ \\ \\ \\ };
    \node[sstrip, below of=h4, node distance=1.1cm] (f4) {};
  \node[strip, right of=f, node distance=3.5cm, yshift=1.35cm, xshift=1.0cm] (h5) { \\ \\ \\ \\ \\ };
  \node[sstrip, below of=h5, node distance=1.1cm] (f5) {};

    \node[fit=(h1)(h2)(h3)(h4)(h5), right of=f, node distance=3.5cm, yshift=1.37cm] (h_f) {};

    \node[brick, below of=h_f, node distance=3.5cm](d1) {};
    \node[brick, xshift=0.6cm, below of=h_f, node distance=3.5cm] (d2) {};
    \node[brick, xshift=-0.6cm, below of=h_f, node distance=3.5cm] (d3) {};
    \node[brick, xshift=1.2cm, below of=h_f, node distance=3.5cm] (d4) {};
    \node[brick, xshift=-1.2cm, below of=h_f, node distance=3.5cm] (d5) {}; 
    \node[sssource=Decoder, fit=(d1)(d2)(d3)(d4)(d5), below of=h_f, node distance=3.5cm] (decoder) {}; 
    \node[concat, above of=d2, node distance=0.5cm] (concat2) {};

    %Mark Text
    \node[above of=d2, node distance=1.3cm] (y_hat) {$\hat{y_4}$};
    \node[above of=h1, node distance=0.0cm, font=\fontsize{2}{1}, xshift=0.01cm, rotate=-90] (h_x1) {$h_{x_3}$};
    \node[above of=h2, node distance=0.0cm, font=\fontsize{2}{1}, xshift=0.01cm, rotate=-90] (h_x2) {$h_{x_2}$};
    \node[above of=h3, node distance=0.0cm, font=\fontsize{2}{1}, xshift=0.01cm, rotate=-90] (h_x3) {$h_{x_4}$};
    \node[above of=h4, node distance=0.0cm, font=\fontsize{2}{1}, xshift=0.01cm, rotate=-90] (h_x4) {$h_{x_1}$};
    \node[above of=h5, node distance=0.0cm, font=\fontsize{2}{1}, xshift=0.01cm, rotate=-90] (h_x5) {$h_{x_5}$};
    
    \node[above of=f1, node distance=-0.0cm, font=\fontsize{6}{1}] (h_x1) {$\mathpzc{f}_{3}$};
    \node[above of=f2, node distance=-0.0cm, font=\fontsize{6}{1}] (h_x2) {$\mathpzc{f}_{2}$};
    \node[above of=f3, node distance=-0.0cm, font=\fontsize{6}{1}] (h_x3) {$\mathpzc{f}_{4}$};
    \node[above of=f4, node distance=-0.0cm, font=\fontsize{6}{1}] (h_x4) {$\mathpzc{f}_{1}$};
    \node[above of=f5, node distance=-0.0cm, font=\fontsize{6}{1}] (h_x5) {$\mathpzc{f}_{5}$};    

%   \node[above of=r_network, node distance=1.7cm] (aaa) {Focus Inference};
%   \node[right of=aaa, node distance=9cm] (bbb) {Attentive Generation};    
   \node[left of=f_generation, node distance=3.7cm, yshift=0.1cm] (ccc) {\small $KL(q_R||p_P)$};
   
%    \node[below of=p5, node distance=1.2cm] (ppp1) {$x_1$};
%    \node[below of=p4, node distance=1.2cm] (ppp5) {$x_5$};
%    \node[below of=p3, node distance=1.2cm] (ppp2) {$x_2$};
%    \node[below of=p2, node distance=1.2cm] (ppp4) {$x_4$};
%    \node[below of=p1, node distance=1.2cm] (ppp3) {$x_3$}; 

%    \node[below of=r5, node distance=1.2cm] (rrr1) {$y_1$};
%    \node[below of=r4, node distance=1.2cm] (rrr5) {$y_5$};
%    \node[below of=r3, node distance=1.2cm] (rrr2) {$y_2$};
%    \node[below of=r2, node distance=1.2cm] (rrr4) {$y_4$};
%    \node[below of=r1, node distance=1.2cm] (rrr3) {$y_3$};	
    
    \node[below of=d5, node distance=1.2cm, font=\fontsize{8}{2}] (ddd1) {$SOS$};
    \node[below of=d4, node distance=1.2cm] (ddd5) {$y_4$};
    \node[below of=d3, node distance=1.2cm] (ddd2) {$y_1$};
    \node[below of=d2, node distance=1.2cm] (ddd4) {$y_3$};
    \node[below of=d1, node distance=1.2cm] (ddd3) {$y_2$};
    
    \\};

  \draw[to, dashed] (r_encoder) -- (h_y);  
  \draw[to, dashed] (h_y) -- (r_network);
  \draw[to, dashed] ($(h_x.east)$) -- ($(r_network.west)$);
  \draw[to, dashed] (r_network) -- (z); 
  \draw[to, bend right, dashed] ($(z.south) + (0.2, 0)$) -- ($(f_generation.west) + (0, 0.05)$); 
  \draw[to, dashed, <->] (z) -- (z_dot);
  %Test
  \draw[to] (h_x) -- (p_network);
  \draw[to] (p_network) -- (z_dot);
  \draw[to, bend left] ($(z_dot.north) + (0.2, 0)$) -- ($(f_generation.west) + (0, 0.05)$);
  %Common
  \draw[to] (p_encoder) -- (h_x);
  \draw[to] (f_generation) -- (f);

  %Decoder Connection
  \draw[sto]($(d5.east)+(0, 0)$) -- ($(d3.west)+(0, 0)$);
  \draw[sto]($(d3.east)+(0, 0)$) -- ($(d1.west)+(0, 0)$);
  \draw[sto]($(d1.east)+(0, 0)$) -- ($(d2.west)+(0, 0)$);
  \draw[sto]($(d2.east)+(0, 0)$) -- ($(d4.west)+(0, 0)$);
%  \draw[sto]($(d5.north)$) -- ($(d5.north) + (0, 0.4)$) node{$y_1$};
%  \draw[sto]($(d3.north)$) -- ($(d3.north) + (0, 0.4)$) node{$y_2$};
%  \draw[sto]($(d1.north)$) -- ($(d1.north) + (0, 0.4)$) node{$y_3$} ;
  \draw[sto]($(d2.north)$) -- ($(d2.north) + (0, 1)$);
%  \draw[sto]($(d4.north)$) -- ($(d4.north) + (0, 0.4)$) node[font=\fontsize{7}{2}\selectfont]{$EOS$};
  \draw[sto]($(d5.south) + (0, -0.8)$)  -- ($(d5.south)$);
  \draw[sto]($(d3.south) + (0, -0.8)$)  -- ($(d3.south)$);
  \draw[sto]($(d1.south) + (0, -0.8)$)  -- ($(d1.south)$);
  \draw[sto]($(d2.south) + (0, -0.8)$)  -- ($(d2.south)$);
  \draw[sto]($(d4.south) + (0, -0.8)$)  -- ($(d4.south)$);  
      
  %p_encoder to f_genration and to f
  \draw[to] ($(h_x.east)$) -- ($(h_x.east) + (0.3, 0)$) |- ($(h_x.east) + (0.3, -1.2)$) |- ($(p_encoder.east) + (11, -1.2)$) -- ($(d5.west) + (-0.5, 0)$) -- ($(d5.west)$);

  \draw($(p_encoder.east) + (8.5, -1.2)$) edge[->, bend right=20, >=stealth',shorten >=1pt,semithick,font=\sffamily\footnotesize] ($(concat.south)$);
  \draw($(p_encoder.east) + (6.4, -1.2)$) edge [->, bend left=20, >=stealth',shorten >=1pt,semithick,font=\sffamily\footnotesize] ($(f_generation.west) + (0, -0.1)$);

%  \draw[line] ($(p_encoder.east) + (6.3, -1.0)$) -- ($(p_encoder.east) + (9.65, -1.0)$);
  
  %concat
  \draw[line] ($(concat.east)$) -- ($(concat.west)$);
  \draw[line] ($(concat.south)$) -- ($(concat.north)$);
  \draw[to] (f) -- (concat);
  \draw[to] (concat) -- ($(concat.east) + (1.5, 0)$) |- ($(h_f.west)$);
  
  %attention
  \draw[to, red] ($(f1.south)$) -- ($(f1.south) + (0, -1.3)$) |- ($(concat2.west)$); 
  \draw[line, red] ($(f2.south)$) -- ($(f2.south) + (0, -0.5)$); 
  \draw[line, red] ($(f3.south)$) -- ($(f3.south) + (0, -0.5)$); 
  \draw[line, red] ($(f4.south)$) -- ($(f4.south) + (0, -0.5)$); 
  \draw[line, red] ($(f5.south)$) -- ($(f5.south) + (0, -0.5)$);
  \draw[line, red] ($(f5.south) + (0, -0.45)$) -- ($(f4.south) + (-0.05, -0.45)$);

%  \draw ($(f3.south) + (-0.5, -0.6)$) node[red] {\tiny Focus-Informed Attention};

  \draw[line] ($(concat2.west)$) -- ($(concat2.east)$);
  \draw[line] ($(concat2.south)$) -- ($(concat2.north) + (0, 0.03)$);

%  %legend
%  \draw[to, red] ($(r_encoder.south) + (-1.5, -4)$) -- ($(r_encoder.south) + (0.5, -4)$);
%  \draw[to, blue] ($(r_encoder.south) + (2.5, -4)$) -- ($(r_encoder.south) + (4.5, -4)$);
%  \draw[to] ($(r_encoder.south) + (6.5, -4)$) -- ($(r_encoder.south) + (8.5, -4)$);
  
%  \draw($(r_encoder.south) + (1, -4)$) node {Train};
%  \draw($(r_encoder.south) + (5, -4)$) node {Test};
%  \draw($(r_encoder.south) + (10.25, -4)$) node {Train and Test Both};
  
  \draw[dots] ($(f_generation.west) + (-0.5, -1.7)$) -- ($(f_generation.west) + (-0.5, 1.6)$);

  \draw[dots] ($(f.east) + (0.23,-1.7)$) -- ($(f.east) + (0.23,1.6)$);
 
%  \draw[dots] ($(h_x5.east) + (0.1,-0.2)$) -- ($(h_x5.east) + (0.1, 1.26)$);
  \draw[dots] ($(h_x4.west) + (-0.1,-0.2)$) -- ($(h_x4.west) + (-0.1, 1.26)$);
  \draw[dots] ($(h_x4.west) + (-0.1, 1.26)$) -- ($(h_x4.west) + (1.35, 1.26)$);
  \draw[dots] ($(h_x4.west) + (-0.1, -0.2)$) -- ($(h_x4.west) + (1.35, -0.2)$);
  \draw[dots] ($(h_x4.west) + (1.35, 1.26)$) -- ($(h_x4.west) + (1.35, -0.2)$);
  \draw($(f.east) + (1.0, 1.1)$) node {$\mathbf{h'_x}$};
  
%  \draw[dots, ->] ($(h_x5.east) + (0.1, 0.53)$) -- ($(h_x5.east) + (0.4, 0.53)$);
%  \draw($(h_x5.east) + (1.0, 1.1)$) node {$\mathbf{h'_x}$};

%% Train Module Absence
%  \draw[dots] ($(r_encoder.west) + (-0.1, -0.75)$) -- ($(r_encoder.west) + (-0.1, 0.6)$);
%  \draw[dots] ($(r_encoder.west) + (-0.1,  0.6)$) -- ($(r_encoder.west) + (4.95, 0.6)$);
%  \draw[dots] ($(r_encoder.west) + (-0.1, -0.75)$) -- ($(r_encoder.west) + (4.95, -0.75)$);
%  \draw[dots] ($(r_encoder.west) + (4.95, -0.75)$) -- ($(r_encoder.west) + (4.95, 0.6)$);
%%  \draw($(r_encoder.north) + (2.0, 0.25)$) node[gray] {\small Absence during testing};
  
  \draw ($(d) + (0, 0)$) node[] {$\mathpzc{D}_t$};
  \draw[line, red] ($(d.east)$) -- ($(d.east)+(0.49, 0)$);

  \draw[to, dashed] ($(f.east) + (6.5,-1.1)$) -- ($(f.east) + (6.5,-0.6)$);
  \draw[to, dashed] ($(f.east) + (6.5, 0.6)$) -- ($(f.east) + (6.5, 0.0)$);
  
  \draw ($(f.east) + (6.5,-0.3)$) node[] {\small $\mathcal{L}_{foc} = ||\frac{1}{\mathbf{|y|}}\mathpzc{D} - \mathpzc{F}||_2$};
  
  \draw ($(p_encoder.south) + (0, -1)$) node[] {$\mathbf{x}=\{x_i\}_{i=1}^{5}$};
  \draw[to] ($(p_encoder.south) + (0, -0.7)$) -- ($(p_encoder.south)$);
  
    \draw ($(r_encoder.south) + (0, -1)$) node[] {$\mathbf{y}=\{y_i\}_{i=1}^{5}$};
  \draw[to, dashed] ($(r_encoder.south) + (0, -0.7)$) -- ($(r_encoder.south)$);

%  %a curly brace
%  \draw[decorate, dashed, gray, decoration={brace,amplitude=10pt,mirror,raise=4pt},yshift=0pt]
%(8.45,-0.12) -- (8.45,2.82) node [black,midway,xshift=0.8cm] {
%$\mathbf{h'_x}$};

%  \draw($(f.east) + (-0.35, 0.6)$) node {\small $\mathpzc{F} = \{\mathpzc{f}_i\}_{i=1}^{5}$};

        \begin{groupplot}[
                group style={
                group size=1 by 2,
                xlabels at=edge bottom,
                ylabels at=edge left,  
                horizontal sep=0.1cm,  
                vertical sep=2.5cm         
                },
                height=3cm,
                width=5cm,
               ]
                    
            %1
             \nextgroupplot[title={\scriptsize focus distribution $\mathpzc{F}$},
             every axis title/.style={at={(0.3,1.3)}},
            xbar,
            xmin=0, xmax=0.7,
            yticklabels={$h_{x_5}$, $h_{x_4}$, $h_{x_3}$, $h_{x_2}$, $h_{x_1}$},
            yticklabel style = {font=\small},
            ytick={1,...,5},
            y axis line style = { opacity = 0 },
            axis x line       = none,
            tickwidth         = 0pt,
            enlarge y limits  = 0.05,
            enlarge x limits  = 0.01,
            /pgf/bar width=2pt,% bar width
             x label style={
        /pgf/number format/.cd,
            fixed,
            fixed zerofill,
            precision=2,
        /tikz/.cd
    },
            every node near coord/.append style={font=\tiny},
            nodes near coords={\pgfmathprintnumber[precision=4]{\pgfplotspointmeta}},
            legend columns=-1,
            legend style={/tikz/every even column/.append style={column sep=0.5cm}},
            legend style={font=\fontsize{8}{20}\selectfont},
            legend image code/.code={
            \draw [#1] (0cm,-0.1cm) rectangle (0.2cm,0.25cm); },
            legend style={at={(0.7,-0.1)},anchor=north},
            at={(0.6\textwidth,0.07\textwidth)},]
            
          \addplot[blue,fill=blue] coordinates {(0.15,1) (0.35,2) (0.2,3) (0.2,4) (0.1,5)};  
%          \addplot coordinates {(0.15,1) (0.17,2) (0.17,3) (0.09,4) (0.18,5) (0.21,6) (0.00,7) (0.00,8) (0.00,9) (0.00,10)};

            %2
             \nextgroupplot[title={\scriptsize attention distribution $\mathpzc{D}$},
             every axis title/.style={at={(0.3,1.3)}},
            xbar,
            xmin=0, xmax=0.7,
            yticklabels={$h_{x_5}$, $h_{x_4}$, $h_{x_3}$, $h_{x_2}$, $h_{x_1}$},
            yticklabel style = {font=\small},
            ytick={1,...,5},
            y axis line style = { opacity = 0 },
            axis x line       = none,
            tickwidth         = 0pt,
            enlarge y limits  = 0.05,
            enlarge x limits  = 0.01,
            /pgf/bar width=2pt,% bar width
             x label style={
        /pgf/number format/.cd,
            fixed,
            fixed zerofill,
            precision=2,
        /tikz/.cd
    },
            every node near coord/.append style={font=\tiny},
            nodes near coords={\pgfmathprintnumber[precision=4]{\pgfplotspointmeta}},]
            
            \addplot[red,fill=red] coordinates {(0.1,1) (0.4,2) (0.2,3) (0.2,4) (0.1,5)};

        \end{groupplot}

\end{tikzpicture}
}
\caption{A framework of our proposed model, where the operation $\oplus$ denotes concatenation, the dashed arrow lines are absent during testing, and the proposed focus-constrained mechanism is represented by the red lines.}

\label{fig:figure1}
\end{figure*}

\section{Related Work}
\label{sec:section2}
The attention mechanism~\cite{bahdanau2014neural,luong2015effective} has become a widely-used component for Seq2Seq~\cite{sutskever2014sequence,cho2014learning} to model Short-Text Conversation~\cite{shang2015neural,vinyals2015neural,sordoni2015neural}. Although promising results have been achieved, attention-based Seq2Seq models still tend to generate generic and trivial responses~\cite{li2016diversity}.

There have been many approaches attempted to address this problem. \citet{li2016diversity} reranked the n-best generated responses based on Maximum Mutual Information (MMI). \citet{shao2017generating} adopted segement-level reranking to encourage diversity during early decoding steps. However, these reranking-based methods only introduce a few variants of decoded words. Another group of researches attempted to encourage diversity by incorporating extra information.  \citet{xing2017topic} injected topic words and \citet{yao2017towards} introduced a cue word based on Point-wise Mutual Information (PMI) into generation models. \citet{ghazvininejad2018knowledge} grounded on knowledge bases to provide factual information for the decoder. %\citet{gao2019generating} assumed a keyword could direct response generation.
However, it is difficult to ensure these external information are always appropriate to the conversation context.

Another line of research introduced a set of latent responding mechanisms and generated responses based on a selected mechanism. \citet{zhou2017mechanism} learned the post-response mappings as a mixture of the mechanisms, but it is questionable that they only relied on one single mechanism when generating responses given a new post. \citet{chen2019generating} adopted posterior selection to build one-to-one mapping relationship between the mechanisms and target responses. Since the target response is missing during testing, it is hard to ensure a satisfactory generated response by a randomly picked mechanism.

Our work centers in the research line of conditional response generation through variational learning~\cite{serban2017hierarchical,zhao2017learning}. However, the variational methods inevitably suffer from bypassing the latent variable and generating disfluent responses. \citet{zhao2017learning} combined CVAE with dialog acts to learn meaningful latent variable, however the discourse-level dialog act is hard to be captured from short conversation. \citet{gu2018dialogwae} introduced Gaussian mixture prior network, but it is hard to determine the number of mixtures and the optimization is complicated. \citet{gao2019discrete} assumed the response generation is driven by a single word, and connected each latent variable with words in the vocabulary. Nevertheless, the difficulty is how to target the driving word for a specific post-response pair. More importantly, all of these methods rely on the coarse-grained discourse-level information, which might be insufficient in leading to a satisfactory response.

Notably, our work induces the response generation with \emph{focus}, a fine-grained feature extracted from the discourse-level latent variable. Compared with the variational attention that models the alignment as latent variable \cite{bahuleyan2018variational,deng2018latent}, we are mainly inspired by the idea of coverage vector~\cite{tu2016modeling} to dynamically adjust the attention based on the attention history and the proposed \emph{focus}. The difference is that \citet{tu2016modeling} addressed the under/over translation problem and the decoder in their work pays equal attention to the source words. In contrast, our work constrains the decoder to align the decoding attention with the fine-grained \emph{focus} to generate diverse responses.

\section{Model}
\label{sec:section3}
\subsection{Preliminaries and Model Overview}
A neural generative model is trained on a collection of post-response pairs $\{(\mathbf{x}, \mathbf{y})\}$, and aimed to generate a response $\mathbf{y}$ word-by-word given an input $\mathbf{x}$. %The length of post is denoted as $\mathbf{|x|}$, while $\mathbf{|y|}$ is length of response $\mathbf{y}$. 
At the basis of our approach is CVAE where a latent variable $z$ is considered to capture discourse-level diversity. To extract fine-grained information, we design \emph{focus} $\mathpzc{F} = \{\mathpzc{f}_i\}_{i=1}^{\mathbf{|x|}}$ over the post words, where $\mathpzc{f}_i$ measures to what extent the latent variable $z$ is concentrating on the post word $x_i$, and $\mathbf{|x|}$ is the length of input $\mathbf{x}$. Besides, we introduce a coverage vector $\mathpzc{D}_t=\{\mathpzc{d}_{i, t}\}_{i=1}^{\mathbf{|x|}}$, where $\mathpzc{d}_{i, t}$ accumulates the attention weights over the post word $x_i$ up until $t$-th decoding step.

Figure~\ref{fig:figure1} illustrates the whole framework of our model consisting of three components: CVAE basis, focus generator and response generator. Based on the CVAE framework, we firstly introduce a probabilistic distribution over the latent variable $z$ to model potential responses for a given $\mathbf{x}$. Then, focus generator produce the \emph{focus} $\mathpzc{F}$ by attending the latent variable $z$ to hidden 
representation $\mathbf{h_x}$ of the input. The obtained $\mathpzc{F}$ is then concatenated with $\mathbf{h_x}$ to obtain $\mathbf{h'_x}$ for word prediction. Specifically, the decoder attentively refers to $\mathbf{h'_x}$ and 
accumulates decoding attention weights through the coverage vector $\mathpzc{D}_t$. To direct response generation using the \emph{focus} $\mathpzc{F}$, we not only optimize the variational lower bound on response generation, but also optimize a regularization term named as focus constraint by minimizing the divergence $\mathpzc{D}$ and $\mathpzc{F}$.

\subsection{Background of CVAE}
Typically, the conditional variational autoencoder (CVAE) introduces a probabilistic distribution over the latent variable to model response diversity. Following CVAE, we firstly encode $\mathbf{x}$ and $\mathbf{y}$ by the post and response encoder, respectively. The two encoders are constructed by the shared bidirectional GRUs~\cite{cho2014learning} which generate a series of hidden states $\{h_{x_i}\}_{i=1}^{|\mathbf{x}|}$ for $\mathbf{x}$ and $\{h_{y_i}\}_{i=1}^{\mathbf{|y|}}$ for $\mathbf{y}$. Then, we obtain the sentence representation $\overline{h_x}$ for the post $\mathbf{x}$ by averaging $\{h_{x_i}\}_{i=1}^{\mathbf{|x|}}$. The sentence representation $\overline{h_y}$ for the response $\mathbf{y}$ is calculated from $\{h_{y_i}\}_{i=1}^{|\mathbf{y}|}$ in the same way.

In training phase, we sample a latent variable $z$ from the posterior distribution $q_R(z|\mathbf{x}, \mathbf{y})$. The distribution is modeled as a multivariate Gaussian distribution $\mathcal{N}(\mu, \Sigma)$, where $\Sigma$ is a diagonal covariance. We parameterize $\mu$ and $\Sigma$ by the recognition network through a fully connected layer conditioned on the concatenation $[\overline{h_x}; \overline{h_y}]$: 

\begin{equation}
\begin{aligned}
\label{eqn:eqn1}
\left[ \begin{array}{ll} \mu \\ \log(\Sigma)  \end{array} \right] = W_q\left[ \begin{array}{ll} \overline{h_x} \\ \overline{h_y}  \end{array} \right] + b_q
\end{aligned}
\end{equation}

\noindent where $W_q$ and $b_q$ are learnable parameters. To mitigate the gap in encoding of latent variables between train and testing ~\cite{sohn2015learning, yan2015attribute2image}, CVAE requires the posterior distribution $q_R(z|\mathbf{x}, \mathbf{y})$ to be close to the prior distribution ${p_{P}(z|\mathbf{x})}$. Notably, ${p_{P}(z|\mathbf{x})}$ is parameterized by the prior network and also follows a multivariate Gaussian distribution $\mathcal{N}(\mu', \Sigma')$ in a similar way but only conditioned on $\overline{h_x}$. As usual, we minimize the discrepancy between the two distributions by the Kullback-Leibler divergence: 
\begin{equation}
\begin{aligned}
\label{eqn:eqn2}
\mathcal{L}_{kl} = KL(q_{R}(z|\mathbf{x}, \mathbf{y}) || p_{P}(z|\mathbf{x})) 
\end{aligned}
\end{equation}

By sampling different $z$, the model is supposed to output semantically different responses. However, such latent variable is too coarse to guide a satisfactory response generation, as discussed previously.

\subsection{Focus Generator}
The core is how to better exploit indicative information from the discourse-level variable for diverse response generation. In this work, we transform the discourse-level latent variable $z$ into fine-grained signal using a focus generator $g(\mathbf{h_x}, z)$ as shown in the middle of Figure~\ref{fig:figure1}. To be specific, the focus generator attends the latent variable $z$ to the post representation $\mathbf{h_x}$, and 
produces the \emph{focus} distribution $\mathpzc{F}=\{\mathpzc{f}\}_{i=1}^{|\mathbf{x}|}$. Similar to the standard attention~\cite{bahdanau2014neural,luong2015effective}, 
the generated \emph{focus} $\mathpzc{F}$ measures the response concentration a specific post word, which is calculated by:

\begin{equation}
\label{eqn:eqn3}
g(\mathbf{h_{x}}, z)= \mathpzc{F} = \{ \frac{\exp(f(h_{x_i}, z))}{\sum_{k=1}^{\mathbf{|x|}}\exp(f(h_{x_k}, z))}\}_{i=1}^{\mathbf{|x|}}
\end{equation}

\noindent where $f(h_{x_i}, z) = v_f^\top\tanh(W_fh_{x_i} + U_fz)$ and $W_f$ and $U_f$ are learnable parameters. This \emph{focus} captures to what extent the response semantics is related to the post words, which will serve as fine-grained signals for the decoder. Notably, the higher the focus, the response is more likely to pay attention to the corresponding word. Compared with the coarse-grained $z$, the word-level \emph{focus} is of great guiding significance when generating responses.

\subsection{Focus-Guided Generation}
\iffalse The remaining is to properly incorporate the fine-grained \emph{focus} into the generation component. Since the proposed \emph{focus} implies the semantics of the target response, we notify the decoder of this information by combining the word-level \emph{focus} with the hidden states. During each decoding step, the \emph{focus} signals are also beneficial as they indicate how much attention should be paid to the current word. 

In other words, the \emph{focus} provides word-level signals indicating whether a word is attended properly. Motivated by this, we develop focus-guided mechanism to facilitate the decoder adjust the attention during the generation. 
\fi

The remaining is to properly incorporate the fine-grained \emph{focus} into response generation. Since the \emph{focus} weights imply the semantics of the target response, they are beneficial signals indicating whether a word is attended properly during decoding.

Concretely, we develop a focus-guided mechanism to facilitate the decoder adjust the attention during the generation. To notify the decoder of the \emph{focus}, we concatenate $\mathbf{h_x}$ and $\mathpzc{F}$ to obtain a series of combined hiddens of the post $\mathbf{h_x'} = \{h'_{x_i}\}_{i=1}^{\mathbf{|x|}}$ (the green and yellow vectors in Figure~\ref{fig:figure1}). After integrating the extra feature $\mathpzc{f}_i$, the devised representations $\mathbf{h'_x}$ are then used to calculate the attention weights. Inspired by \citet{tu2016modeling}, we borrow the idea of coverage attention and introduce the coverage vector $\mathpzc{D}_t =\{\mathpzc{d}_{i, t}\}_{i=1}^{\mathbf{|x|}}$ that records the attention history, where $\mathpzc{d}_{i,t} = \sum_{k=1}^{t}\alpha_{i, k}$ accumulates the decoding attention weights on the post word $x_i$. Here, $\alpha_{i,t}$ stands for the attention weight on the post word $x_i$ at ${t}$-th decoding step ($t\in[1, \mathbf{|y|}]$), which is calculated as:

\begin{equation}
\label{eqn:eqn4}
\alpha_{i,t} = \frac{\exp(e_{i, t})}{\sum_{k=1}^{\mathbf{|x|}}\exp(e_{k, t})}
\end{equation}

\begin{equation}
\label{eqn:eqn5}
e_{i,t} = v_a^\top\tanh(W_{a}h'_{x_i} + U_a s_{t-1} + V_a\mathbf{a}_{t-1})
\end{equation}

\begin{equation}
\label{eqn:eqn6}
\mathbf{a}_{t-1}=\sum_{i=1}^{\mathbf{|x|}}\mathpzc{d}_{i, t-1}h'_{x_i}
\end{equation}

\noindent where $s_{t-1}$ is decoder's hidden state at $(t-1)$-th step, and $\mathbf{a}_{t-1}$ takes into account the attention history before $t$-th step. At the end of each decoding step, a predicted word $\hat{y_t}$ is obtained by:

\begin{equation}
\label{eqn:eqn7}
\hat{y_t} = \text{softmax}(W_d[s_{t}; \sum_{i=1}^{\mathbf{|x|}}\alpha_{i,t}h_{x_i}] + d_d)
\end{equation}

\noindent where $W_d$ and $d_d$ are learnable parameters. Since the \emph{focus} suggests how much attention should be paid to during each decoding step, the devised focus-guided mechanism is able to globally determine a word based on the attention history as well as the current state.

\subsection{Focus Constraint}
Nevertheless, one potential drawback is that the decoder could still ignore the \emph{focus} signals even equipped with the focus-guided mechanism. Considering that \emph{focus} measures the response's significance on a specific post word, it is also essential for the decoder to concentrate on the word with higher \emph{focus} weight, and vice versa. 

To prevent the decoder bypassing the \emph{focus} signal, we design a focus constraint to regulate the learning of post-response pairs by taking into account the \emph{focus} weights. As shown in the right side of Figure~\ref{fig:figure1}, the focus constraint requires the model to minimize the discrepancy between the focus weight distribution $\mathpzc{F}$ and decoding attention distribution $\mathpzc{D}$. To implement it, we define the focus constraint $\mathcal{L}_{foc}$ as the Euclidean norm distance between $\mathpzc{D}$ and $\mathpzc{F}$:

\begin{equation}
\label{eqn:eqn8}
\mathcal{L}_{foc} = ||\frac{1}{\mathbf{|y|}}\mathpzc{D} - \mathpzc{F}||_2
\end{equation}

\noindent where $\mathpzc{D}$ sums up all the decoding attention weights over the post words and $\mathbf{|y|}$ is the total number of decoding steps. Considering $\sum_{i=1}^{\mathbf{|x|}}\mathpzc{f}_i=1$, a division of $\mathbf{|y|}$ from $\mathpzc{D}$ makes the two terms being compared at the same magnitude. We name this constrained decoding attention as \textbf{Focus-Constrained Attention Mechanism}. Such a constraint will make the decoder draw attention by globally consulting the \emph{focus} $\mathpzc{F}$ and distribute the attention dynamically. For example, given a distribution $\mathpzc{F}$, if the hidden output $h_{x_i}$ has been attended to a certain degree $\mathpzc{d}_{i, t-1}\approx\mathpzc{f}_i$, the model will discourage the decoder to overly emphasize on $h_{x_i}$ after the $t$-th step. In contrast, if the hidden output $h_{x_i}$ has been hardly attended compared with its \emph{focus} weight ($\mathpzc{d}_{i, t-1}\ll{\mathpzc{f}_i}$), the model will encourage the decoder to pay more attention onto $h_{x_i}$ afterwards. 

\subsection{Optimization and Testing}
\label{sec:section3.4}
Overall, all the modules described above are jointly trained in an end-to-end way by minimizing the total loss:  

\begin{equation}
\label{eqn:eqn9}
\mathcal{L}_{total} = \mathcal{L}_{seq} + \mathcal{L}_{foc} + \gamma \mathcal{L}_{kl} + \mathcal{L}_{bow}
\end{equation}

Here, $\mathcal{L}_{seq}$ is the sequence cross entropy between the generated response ${\mathbf{\hat{y}}}$ and the corresponding ground truth ${\mathbf{y}}$. $\mathcal{L}_{foc}$ is the proposed focus constraint as described above. To address the problem of vanishing latent variable, we follow~\citet{bowman2015generating} and adopt the annealing weight $\gamma$ for KL divergence $\mathcal{L}_{kl}$, where $\gamma$ is gradually increased during training phase. We also employ the auxiliary bag-of-word loss $\mathcal{L}_{bow}$ to further alleviate the vanishing issue~\cite{zhao2017learning}. 

At testing phase, an intermediate \emph{focus} $\mathpzc{F}$ will be obtained with the prior network and focus generator. Notably, this enables us to generate diverse responses by sampling multiple latent variables from the prior network, where each sampled $z$ leads to a semantically distinct response.

%%Table 2
%%%evaluation metric table 
\begin{table*}[t]
\centering
\smallskip
\resizebox{1.0\textwidth}{!}{ 
% If your table exceeds the column or page width, use this command to reduce it slightly
\begin{tabular}{l|c|c|c|c|c|c|c|c|c}
\toprule
\multirow{2}{*}{\textbf{Method}} &
\multicolumn{2}{c|}{\textbf{Multi-BLEU}} &
\multicolumn{2}{c|}{\textbf{Intra-Dist}} &
\multicolumn{2}{c|}{\textbf{Inter-Dist}} &
\multicolumn{2}{c|}{\textbf{Quality}} &
\multirow{2}{*}{\textbf{Diversity}} \\ 
\cline{2-9}
& BLEU-1 & BLEU-2 & Dist-1 & Dist-2 & Dist-1 & Dist-2 & Acceptable & Good \\
\midrule
\textbf{S2S} & 26.63 & 9.07 & 45.90 & 60.12 & 9.75 & 34.31 & 61.33 & 2.88 & 1.66\\
\textbf{MMI} & 26.67 & 9.08 & 46.17 & 60.49 & 9.74 & 34.43 & 60.22 & 3.33 & 1.68 \\
\textbf{MARM} & 26.70 & 9.30 & 47.00 & 60.90 & 10.92 & 37.88 & 61.77 & 4.22 & 1.72\\
\textbf{CMHAM} & 26.18 & 7.58 & 60.28 & 76.36 & 5.83 & 26.61 & 59.33 & 2.88 & 2.26\\
\textbf{CVAE} & 28.88 & 8.78 & 75.57 & 92.66 & 13.59 & 49.68 & 36.88 & 2.88 & 2.62\\
%\textbf{RLKW} & \textbf{32.37} & \textbf{11.26} & 58.61 & 74.60 & 7.37 & 34.70 & 63.55 & 8.00 & 1.88\\
\textbf{DCVAE} & \textbf{30.44} & 8.98 & 73.33 & 90.45 & 14.43 & 53.28 & 58.67 & 4.67 & 2.72\\
\midrule
\textbf{Ours-Foc} & 27.12 & 9.01 & 44.68 & 59.75 & 10.02 & 35.21 & 60.67 & 4.67 & 1.26 \\
\textbf{Ours-FocCoverage} & 29.50 & 9.11 & 66.71 & 85.17 & 15.25 & 54.16 & 62.66 & 3.33 & 2.36 \\
\textbf{Ours-FocConstrain} & 30.32 & \textbf{9.39} & \textbf{80.24} & \textbf{95.53} & \textbf{16.89} & \textbf{59.67} & \textbf{65.33} & \textbf{9.33} & \textbf{2.82}\\
\bottomrule
\end{tabular}
}
\caption{The results from automatic and human evaluations. The Kappa score is 0.45 and 0.70 for \textit{quality} and \textit{diversity} labeling}
\label{tab:table1}
\end{table*}

\section{Experiment}
\label{sec:section4}

\subsection{Dataset}
We conduct experiments on the Weibo benchmark\footnote{https://www.weibo.com/}~\cite{shang2015neural}, a single-round conversational dataset where a post is associated with multiple responses. We follow the default preprocessing step, and obtain 205,164 unique posts and 4,142,299 training post-response pairs in total. After random spilt, we acquire 101,794 post-response pairs for evaluation, and 1,000 distinct posts for testing. Here, each testing post has 5 reference responses for evaluation.

\subsection{Implementation Details}
We implement our model with Tensorflow and run it on NVIDIA Telsa V100. Specifically, the vocabulary size is 50,003 including PAD, UNK and EOS. The word embedding size is 720 as same as the size of latent variable. We build two-layer GRUs for the two parameter-shared encoders as well as for the decoder. In all, our model contains around 130M parameters, which are all randomly initialized with a uniform distribution $[-1, 1]$. We train our model with a batch size of 1,024 by Adam optimizer \cite{kingma2014adam}. We increase the learning rate from 0 up to 0.0008 within the first 8,000 warmup steps and proportionally decrease it to the inverse square root of step number~\cite{vaswani2017attention}.%\footnote{Our code is submitted along this manuscript.}

\subsection{Baseline Models}
\label{sec:section4.3}
To demonstrate the necessity and effectiveness of our proposed mechanism alone, we build it on Seq2Seq and exclude as many other interferences as possible when comparing with the following state-of-the-art baseline models:

\noindent\textbf{S2S} \cite{bahdanau2014neural}: It trains a Seq2Seq model with the standard attention and adopts beam search decoding to generate responses.

\noindent\textbf{MMI} \cite{li2016diversity}: It is a backward Seq2Seq trained from response to post, and reranks the beam searched candidates under MMI criterion.

\noindent\textbf{MARM} \cite{zhou2017mechanism}: It is a Seq2Seq model which additionally contains a diverter that consists of 5 latent responding mechanisms. During training, these mechanisms are learned as a mixture by the weighted average.

\noindent\textbf{CMHAM} \cite{tao2018get}: It is a Seq2Seq model, which is augmented with Constrained-Multi-Head-Attention-Mechanism. The attention heads are constrained by orthogonality and each of them is expected to attend a certain aspect of the post. We set the head number as 5.

\noindent\textbf{CVAE} \cite{zhao2017learning}: It is a vanilla CVAE Seq2Seq trained along with the bag-of-word loss. During testing phase, we take 3 samplings from the prior network to generate each response.

%\textbf{RLKW} \cite{gao2019generating}: It is a reinforced Seq2Seq model, which additionally has keyword predictor and assumes the keyword could direct the response generation. We adopt F1 score as the reward by comparing overlaps between the generated response and a bag of ground-truth responses. 

\noindent\textbf{DCVAE} \cite{gao2019discrete}: It is a CVAE-based Seq2Seq model trained with discrete latent variables, where the latent variables are connected with words in the vocabulary. To follow their paper, we use their original implementation and pre-train the model with extracted keywords. During testing phase, we adopt their two-stage sampling strategy to generate each response.

\noindent\textbf{Ours}: In addition, we implement two variants of our proposed model \textbf{Ours-FocConstrain}, i.e., 1) \textbf{Ours-Foc} introduces the \emph{focus} $\mathpzc{F}$, but it does not incorporate the coverage vector $\mathpzc{D}_t$, and the decoding attention at each step is calculated with only the first two terms in Equation~\ref{eqn:eqn5}. 2) \textbf{Ours-FocCoverage} involves both of the \emph{focus} $\mathpzc{F}$ and the coverage vector $\mathpzc{D}_t$, where the only difference from \textbf{Ours-FocConstrain} is that it is optimized without the focus constraint $\mathcal{L}_{foc}$ in Equation~\ref{eqn:eqn9}.

\subsection{Evaluation Metrics}
\label{sec:section4.4}
All models are required to generate 3 responses and are evaluated using both automatic metrics and human judgements:

\noindent{\textbf{Multi-BLEU}}: \textbf{BLEU}~\cite{papineni2002bleu}\footnote{http://www.nltk.org/py-modindex.html} is a common automatic metric to evaluate the response quality. It measures word overlaps between the generated responses and references. We report \textbf{Multi-BLEU} scores where each generated response is compared with 5 references.

\noindent{\textbf{Dist-1/2}}: \textbf{Dist-1/2} measures the diversity of generated responses by counting the distinct uni-grams and bi-grams \cite{li2016diversity}. In our setting, both \textbf{Intra-Dist} and \textbf{Inter-Dist} are evaluated on the results to calculate \textbf{Dist} of responses for a post and the whole testing set, respectively.

\noindent{\textbf{Human Labeling}}: Since there is a gap between automatic metrics and human annotation~\cite{liu2016not}, we also consider human labeling to further validate the experiment results. We randomly sample 150 posts and generate 3 responses by each method. Then, we ask 3 professional annotators to label the responses from the aspects of \textbf{Quality} and \textbf{Diversity}, respectively.

\noindent{\textbf{Quality}}: We examine the generated responses from the aspects of informativeness (which measures whether the generated response is informative and interesting), relevance (which measures whether the generated response is relevant to the input post) and fluency (which measures whether the quality of the generated response). Each generated response will be categorized into bad, normal or good (scaled as 0, 1, 2). Note that a generated response will be labled as bad, if it is irrelevant to the post or has grammar mistakes. Besides, a good generated response is more than just fluent but also informative compared with a normal one. We report acceptable ratio for responses that are labeled as 1 or 2, and good ratio only for responses that just are labeled as 2.

\noindent{\textbf{Diversity}}: It measures the number of semantically distinct generated responses for a post. The higher the better, the maximum scale is 3.

\section{Results and Analysis}
%%Table 3
%Cases of each method
\begin{table*}[t]
\centering
\smallskip
\newcommand{\tabincell}[2]{\begin{tabular}{@{}#1@{}}#2\end{tabular}}
\resizebox{1.0\textwidth}{!}{ 
% If your table exceeds the column or page width, use this command to reduce it slightly
\begin{tabular}{l|l|l|l}
\toprule
\multicolumn{4}{c}{Post: \tabincell{l}{炎热的夏天想吃遍所有口味的冰淇淋 \\ 
 I want to eat ice-creams of all flavors of in the hot summer} \qquad Gold:  \tabincell{l}{我 现在 特别 想 吃 薰衣草 冰淇淋 \\ I really want to eat lavender ice cream right now }   }\\
\midrule
\textbf{S2S} & \tabincell{l}{太可爱了，好想吃 \\ 
It’s so cute, I want to eat it.} & \tabincell{l}{好想吃。好想吃 \\  
I want to eat it. I want to eat it} & \tabincell{l}{好想吃啊。好想吃 \\ I want to eat it. I want to eat it}  \\
\midrule
\textbf{MMI} & \tabincell{l}{好想吃啊，好想吃。 \\ 
I want to eat it. I want to eat it.} & \tabincell{l}{好想吃啊，好想吃 \\ 
I want to eat it. I want to eat it.} & \tabincell{l}{好想吃啊 … … \\ I want to eat it … … }   \\
\midrule
\textbf{MARM} & \tabincell{l}{这是什么口味的冰淇淋\\ 
What flavor is this ice-cream} &  \tabincell{l}{这是什么口味的冰淇淋 \\ 
What flavor is this ice-cream} & \tabincell{l}{这是什么口味的冰淇淋啊\\
What flavor is this ice-cream} \\
\midrule
\textbf{CMHAM} & \tabincell{l}{冰淇淋也可以吃冰淇淋 \\ Ice-cream can also eat ice-cream.}  & \tabincell{l}{冰淇淋口味太重了吧\\ 
Ice-creams taste too heavy} & \tabincell{l}{好 想 吃 \\ I want it.} \\
\midrule
\textbf{CVAE} & \tabincell{l}{我好久没吃蛋糕了 \\ I haven't eaten cake for a long time}  & \tabincell{l}{夏天快来吧，我的夏天。\\ Come on summer, my summer.} & \tabincell{l}{冰淇淋吃冰淇淋，哈哈。\\ Ice-cream eats ice-cream, haha.} \\
\midrule
%\textbf{RLKW} & \tabincell{l}{我也想吃，口水都流出来了！\\ I want to eat it too. My mouth is watering! }& \tabincell{l}{我就想吃，口水都流出来了\\ 
%I just want to eat it. My mouth is watering }&  \tabincell{l}{我也想吃。我最爱冰淇淋。\\ I would like to eat that too. I love ice cream best. } \\
\textbf{DCVAE} & \tabincell{l}{我也好想吃这个！\\
I really want to eat this too! }& \tabincell{l}{夏天的夏天快到了\\ Summer' summer is comming }&  \tabincell{l}{好想吃啊，流口水了。\\I really want to eat. My mouth is watering.

 } \\
\toprule
\textbf{Ours-Foc} & \tabincell{l}{哇哦。好想吃哦。 \\ Wow. I really want to eat it.}  & \tabincell{l}{ 哇哦。好想吃哦。 \\ Wow. I really want to eat it.} & \tabincell{l}{哇哦。好想吃哦。 \\ Wow. I really want to eat it.} \\
\midrule
\textbf{Ours-FocCoverage} & \tabincell{l}{真心不喜欢冰淇淋 \\ I really don't like ice cream}  & \tabincell{l}{夏天都吃冰淇淋了。\\ Always have ice cream in summer.} & \tabincell{l}{夏天吃了这东东…\\ In summer, eat this thing...} \\
\midrule
\textbf{Ours-FocConstrain} & \tabincell{l}{薄荷味真的很好吃！ \\ 
Mint flavor is really delicious! } & \tabincell{l}{爱吃冰激凌的人表示羡慕嫉妒恨。\\ People who love to eat ice cream are green with envy.} & \tabincell{l}{吃货一枚，鉴定完毕。 \\ A foodie, the identification is done.}\\
\bottomrule
\end{tabular}
}
\caption{The gold and generated responses by each method. }
\label{tab:table2}
\end{table*}

\subsection{Comparison Against Baselines}
Results of automatic metrics and human labelings are shown in Table~\ref{tab:table2}. The Kappa score is 0.45 and 0.70 for \textit{quality} and \textit{diversity} labeling, indicating that the annotators share a satisfactory agreement in the labeling. %It is worth noting that RLKW achieves the best \textit{BLEU}, however as stated in Section~\ref{sec:section4.4} that we consider human labeling as our major judegement, since \textit{BLEU} only measures word-level overlaps and a generated response that has few overlaps with references might still be a good one.

We firstly examine the significance of latent variable. Generally speaking, the compared models without any latent variable (the first 4 rows in Table~\ref{tab:table1}) perform the worst. As shown in Table~\ref{tab:table1}, S2S and MMI achieve the lowest scores. Comparing the 3 generated responses by S2S and MMI shown in Table~\ref{tab:table2} (the 3 columns), they share similar semantics with only a few word variants. As MMI has to rerank the candidates generated by S2S, their performances are similarly disappointing. This result supports that Seq2Seq is limited in modeling diverse responses for a given post even combined with the reranking strategy. Moreover, MARM performs similarly with S2S and MMI in terms of the automatic scores, human judgments as well as the generated responses shown in Table~\ref{tab:table2}. Despite that MARM introduces a set of latent embeddings, its poor performance is attributed to the lack of extra disentanglement control on the mixture learning of latent mechanisms, as analyzed in the previous section. %As a result, these embeddings result in similar generated responses. 
Things become interesting when we examine the performance of CMHAM. It seems that CMHAM effectively improves the diversity over other Seq2Seq models if we only checked the indicators in Table~\ref{tab:table1}. However, the responses generated by CMHAM from Table~\ref{tab:table2} are either too short or ungrammatical. Such inconsistency between the results from Table~\ref{tab:table1} and Table~\ref{tab:table2} might be resulted from several causes. We conjecture one primary reason is the gap between model training and testing. During training, the semantic representation in CMHAM is learned as a mixture of all attention heads. While during testing, CMHAM is limited to use one single constrained head to focus on a certain post word.  

We then examine the variational models equipped with latent variable (the fourth to sixth rows) to investigate which method(s) are more effective in utilizing the latent information. From Table~\ref{tab:table1}, CVAE brings obvious improvements on \textit{Dist} and \textit{Diversity} as compared with the non-variational models (the first four rows). However, the responses generated by CVAE in Table~\ref{tab:table2} are of low quality. It is because that the vanilla CVAE has no extra control on the latent variable, and the stochasticity injected in the latent variable is too overwhelming for the decoder when generating responses. In turn, hardly the decoder is able to balance the latent semantics with the response fluency. As a result, the latent variable fails to effectively direct a high-quality response generation. When comparing DCVAE with CVAE, we can see noticeable increases especially on \textit{Quality} and \textit{Diversity}. This is not surprising in that DCVAE introduces additional control on each latent variable and connects the variables with the words in the vocabulary. Though it is more meaningful to incorporate the latent variable in this way, DCVAE is still insufficient. Take the 2nd generated response from DCVAE in Table~\ref{tab:table2} as an example where the driving word is ``夏天(summer)''. In this case, DCVAE is unable to adjust the attention, and thus directs the flawed response to overly emphasize on ``夏天(summer)''. This example partially proves that even though DCVAE has taken control over the latent variable, it is still problematic to guide response generation through a coarse-grained signal. 

On the contrary, the proposed model and its variants \textbf{Ours-FocCoverage} \textbf{Ours-FocConstrain} base on the fine-grained \emph{focus} signal and successfully improve the overall generation quality as well as response diversity. Especially, our full model \textbf{Ours-FocConstrain} performs the best in terms of almost every metric except \textit{BLEU-1}. The highest scores of human evaluations in Table~\ref{tab:table1} and the responses in Table~\ref{tab:table2} together show that our proposed method \textbf{Ours-FocConstrain} is able to generate high-quality and diverse responses. In brief, our model introduces a performance boost as it fully leverages the word-level information for response generation.

\begin{figure}[hbt!]
\newcommand{\tabincell}[2]{\begin{tabular}{@{}#1@{}}#2\end{tabular}}
\resizebox{0.95\columnwidth}{!}{
\begin{tikzpicture}
        \begin{groupplot}[
                group style={
                group size=3 by 1,
                xlabels at=edge bottom,
                ylabels at=edge left,  
                horizontal sep=4cm              
                }]
            %1
            \nextgroupplot[
            xbar,
            xmin=0, xmax=0.3,
            yticklabels={冰淇淋(ice-cream), 的, 口味(flavor), 所有(all), 遍, 吃(eat), 想(want), 夏天(summer), 的, 炎热(hot)},
            yticklabel style = {font=\small},
            ytick={1,...,10},
            y axis line style = { opacity = 0 },
            axis x line       = none,
            tickwidth         = 0pt,
            enlarge y limits  = 0.05,
            enlarge x limits  = 0.01,
            /pgf/bar width=2pt,% bar width
            legend entries={Response\#1, Response\#2, Response\#3},
            legend style={font=\fontsize{8}{20}\selectfont},
             x label style={
        /pgf/number format/.cd,
            fixed,
            fixed zerofill,
            precision=2,
        /tikz/.cd
    },
            every node near coord/.append style={font=\tiny},
            nodes near coords={\pgfmathprintnumber[precision=4]{\pgfplotspointmeta}},]
%          \addplot coordinates {(0.11,1) (0.05,2) (0.13,3) (0.08,4) (0.09,5) (0.04,6) (0.04, 7) (0.05,8) (0.04,9) (0.32,10)};
          %f1
          \addplot coordinates {(0.11,1) (0.11,2) (0.19,3) (0.08,4) (0.08,5) (0.04,6) (0.09,7) (0.04,8) (0.10,9) (0.12,10)};
          %f2
          \addplot coordinates {(0.10,1) (0.11,2) (0.14,3) (0.11,4) (0.11,5) (0.04,6) (0.12,7) (0.03,8) (0.10,9) (0.08,10)};
          %f3
          \addplot coordinates {(0.14,1) (0.12,2) (0.21,3) (0.09,4) (0.05,5) (0.05,6) (0.10,7) (0.05,8) (0.10,9) (0.05,10)};

       \end{groupplot}

\end{tikzpicture}
}
\caption{The focus distributions of the 3 test cases by Ours-Foc from Table~\ref{tab:table2}}
\label{fig:figure2}
\end{figure}
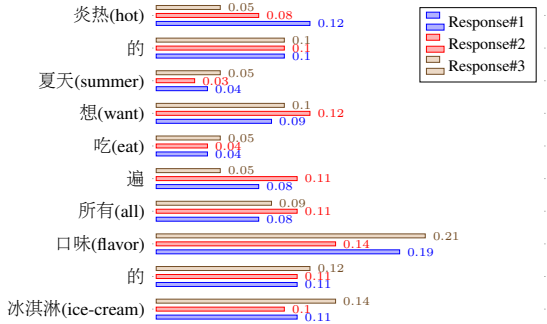

\begin{figure}[hbt!]
\newcommand{\tabincell}[2]{\begin{tabular}{@{}#1@{}}#2\end{tabular}}
\resizebox{1.3\columnwidth}{!}{
\begin{tikzpicture}
        \begin{groupplot}[
                group style={
                group size=2 by 1,
                xlabels at=edge bottom,
                ylabels at=edge left,  
                horizontal sep=1.5cm              
                }]
                    
            %2   
             \nextgroupplot[title={\small \tabincell{c}{Response\#1 by \\ Ours-FocCoverage}: \tabincell{l}{真心不喜欢冰淇淋 \\ I really don't like ice cream}},
             every axis title/.style={at={(0.05,1.1)}},
            xbar,
            xmin=0, xmax=0.7,
            yticklabels={冰淇淋(ice-cream), 的, 口味(flavor), 所有(all), 遍, 吃(eat), 想(want), 夏天(summer), 的, 炎热(hot)},
            yticklabel style = {font=\small},
            ytick={1,...,10},
            y axis line style = { opacity = 0 },
            axis x line       = none,
            tickwidth         = 0pt,
            enlarge y limits  = 0.05,
            enlarge x limits  = 0.01,
            /pgf/bar width=2pt,% bar width
             x label style={
        /pgf/number format/.cd,
            fixed,
            fixed zerofill,
            precision=2,
        /tikz/.cd
    },
            every node near coord/.append style={font=\tiny},
            nodes near coords={\pgfmathprintnumber[precision=4]{\pgfplotspointmeta}},
            legend entries={Attention, Focus},
            legend style={font=\fontsize{8}{20}\selectfont},
%            legend image code/.code={
%            \draw [#1] (0cm,-0.1cm) rectangle (0.2cm,0.25cm); },
            legend style={at={(0.7,0.6)},anchor=north}]
            
          \addplot coordinates {(0.30,1) (0.03,2) (0.12,3) (0.16,4) (0.03,5) (0.05,6) (0.06,7) (0.05,8) (0.05,9) (0.10,10)};  
          \addplot coordinates {(0.15,1) (0.17,2) (0.17,3) (0.09,4) (0.18,5) (0.21,6) (0.00,7) (0.00,8) (0.00,9) (0.00,10)};

            %3      
             \nextgroupplot[title={\small \tabincell{c}{Response\#1 by \\ Ours-FocConstrain}: \tabincell{l}{薄荷味真的很好吃！\\ Mint flavor is really delicious!}},
             every axis title/.style={at={(-0.05,1.1)}},
            xbar,
            xmin=0, xmax=0.7,
            yticklabels={冰淇淋(ice-cream), 的, 口味(flavor), 所有(all), 遍, 吃(eat), 想(want), 夏天(summer), 的, 炎热(hot)},
            yticklabel style = {font=\small},
            ytick={1,...,10},
            y axis line style = { opacity = 0 },
            axis x line       = none,
            tickwidth         = 0pt,
            enlarge y limits  = 0.05,
            enlarge x limits  = 0.01,
            /pgf/bar width=2pt,% bar width
             x label style={
        /pgf/number format/.cd,
            fixed,
            fixed zerofill,
            precision=2,
        /tikz/.cd
    },
            every node near coord/.append style={font=\tiny},
            nodes near coords={\pgfmathprintnumber[precision=4]{\pgfplotspointmeta}},]
            
            \addplot coordinates {(0.11,1) (0.12,2) (0.18,3) (0.03,4) (0.07,5) (0.07,6) (0.13,7) (0.12,8) (0.12,9) (0.03,10)};
       
          \addplot coordinates {(0.13,1) (0.11,2) (0.14,3) (0.08,4) (0.09,5) (0.10,6) (0.11,7) (0.08,8) (0.07,9) (0.06,10)};
 
        \end{groupplot}

%\draw (7,-0.9) node[] {Post: \tabincell{l}{炎热\ 的\ 夏天\ 想\ 吃\ 遍\ 所有\ 口味\ 的\ 冰淇淋 \\ 
% I want to eat ice-creams of all flavors of in the hot summer   } };

\end{tikzpicture}
}
\caption{The focus and attention distribution of the the test cases by Ours-FocCoverage and Ours-FocConstrain from Table~\ref{tab:table2}.}
\label{fig:figure3}
\end{figure}
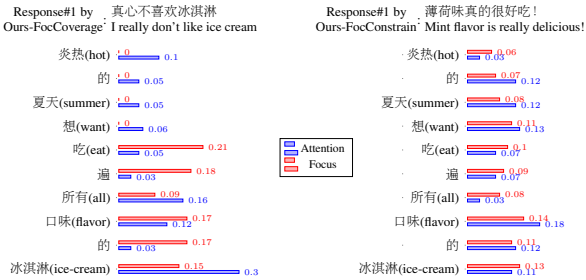

\subsection{Ablation Study and Analysis}
To verify the effectiveness of each proposal in our work, we conduct ablation studies by comparing with several variants of our model, \textbf{Ours-Foc}, \textbf{Ours-FocCoverage} and \textbf{Ours-FocConstrain}. The ablation results are summarized in the last three rows in Table~\ref{tab:table1} and Table~\ref{tab:table2}.

Clearly, the performance gap among our variants indicate that all the three modules in the proposed model are of great importance. One thing to note in Table~\ref{tab:table1} is the unsatisfactory performance achieved by the bare-bone variant \textbf{Ours-Foc} that it performs similarly with the vanilla Seq2Seq. \textbf{Ours-Foc} solely contains the focus generator which dissembles the discourse-level latent variable into word-level guiding signals---\emph{focus}---for each decoding step. This setting is insufficient because the model is prone to bypass the guiding signals. We observe such unexpected phenomenon in Table~\ref{tab:table2} where the three responses from {Ours-Foc} are generated with one single model and thus they are similar to each other. This phenomenon is further validated in Figure~\ref{fig:figure2}, where we plot the \emph{focus} distributions that are correlated with the three responses from Table~\ref{tab:table2}. From this experiment, we can see that the generated responses do not attach much attention to the word ``口味(flavor)" even though the word is assigned with the highest \emph{focus} weight. This verifies that, despite that \textbf{Ours-Foc} incorporates the fine-grained \emph{focus}, it still lacks mechanism(s) and strategy(s) to make full use of it.

Upon the bare-bone model, \textbf{Ours-FocCoverage} incorporates the proposed focus-guided mechanism and increases the metric scores a lot especially on the metric \textit{Dist} and \textit{Diversity}. We attribute this increase to the use of coverage vector. In such way, the model is able to adjust attention based on attention history as well as the \emph{focus}, rather than simply considering the current relevant words as in the standard attention mechanism. Therefore, the \emph{focus} tends to show guiding significance for the decoder to generate qualified responses. From Table~\ref{tab:table2} we can see, the responses generated by \textbf{Ours-FocCoverage} differ from each other with respect to both semantic meaning and their expressions.

More importantly, \textbf{Ours-FocConstrain} further employs the novel focus constraint to properly align the target response with input post according to the \emph{focus}. To examine in detail, we plot both \emph{focus} and decoding attention distribution of the test cases by \textbf{Ours-FocCoverage} and \textbf{Ours-FocConstrain}. As shown in Figure~\ref{fig:figure3}, the latent variable of \textbf{Ours-FocCoverage} addresses the highest \emph{focus} to the word ``吃(eat)". However, the decoder does not follow such guidance and pays more attention to the word ``冰淇淋(ice-cream)", resulting in an improper response. In contrast, the latent variable in \textbf{Ours-FocConstrain} concentrates more on the word ``口味(flavor)" than the others. With the help of focus constraint, the decoder of \textbf{Ours-FocConstrain} makes it to direct the generated response embody the meaning of ``口味(flavor)". In other words, though \textbf{Ours-FocCoverage} introduces the coverage vector and potentially encourages the diversity using different sampled latent variables, \textbf{Ours-FocConstrain} steps further and kills the chance of generating responses regardless of the \emph{focus} by using the constraint $\mathcal{L}_{foc}$. Drawing on the highest scores achieved by \textbf{Ours-FocConstrain}, we conclude that the proposed focus constraint is an indispensable design and is potentially beneficial for CAVE-based response generation models. 

Overall speaking, the proposed \textbf{Focus-Constrained Attention Mechanism} consists of: (1) focus generator to produce fine-grained signals; and (2) focus-guided mechanism and focus constraint to fully utilize the signal. This ablation study validates the necessity of fine-grained latent information, and demonstrates the effectiveness of each component in the proposed method. By leveraging the proposed Focus-Constrained Attention Mechanism, the decoder is able to tell the importance of each word and start a holistic-planned response generation under the fine-grained focus guidance.

%Overall speaking, this ablation study validates the necessity of exploiting fine-grained latent information for CVAE-based response generation models, and demonstrates the effectiveness of each component in the proposed method.

\section{Conclusion}
In this paper, %we propose to improve the response diversity based on variational models. 
we identify the insufficiency of discourse-level latent variable in response generation. To address this, we develop a novel CVAE-based model, which exploits a fine-grained word-level feature to generate diverse responses. On a real-world benchmarking dataset, we demonstrate that our proposed model is able to fully leverage the fine-grained feature, and generate better responses as compared to several SOTA models. Based on the ablation studies, we verify the contribution of each proposal in our method and highlight the significance of fine-grained signal in response generation.

\section*{Acknowledgments}                                                                                                                                                                                  
We would like to thank anonymous reviewers for their helpful comments and suggestions to improve our work.

\bibliographystyle{acl_natbib}
\bibliography{emnlp2020}

\end{CJK*}
\end{document}